**ChatGPT-4 with Code Interpreter can be used to solve introductory college-level vector calculus and electromagnetism problems**

Tanuj Kumar (tanuj.kumar@wisc.edu) and Mikhail A. Kats (mkats@wisc.edu)

Department of Electrical and Computer Engineering, University of Wisconsin-Madison

**Executive summary:**

We evaluated three modes of ChatGPT — 3.5, 4, and 4 with Code Interpreter — on a set of college-level engineering-math and electromagnetism problems, such as those often given to sophomore electrical engineering majors. We selected a set of 13 problems without first testing them with ChatGPT, and had ChatGPT solve them multiple times, using a fresh instance (chat) of ChatGPT each time. The problems range from elementary to medium-level. All data presented here was acquired with the Aug 3, 2023 version of ChatGPT (not the API), which can accept as input text and data, but not natively interpret images[i]; therefore, we avoided problems that required the reader to look at a figure, or rewrote the problems such that were fully described by text. We were strict in our evaluation of ChatGPT's performance, marking a solution as incorrect if even a small part of the solution was wrong.

Our major conclusions are:

- ChatGPT-4 with Code Interpreter (ChatGPT-4/CI), recently renamed Advanced Data Analysis, was able to satisfactorily solve most problems we tested most of the time. Qualitatively, one could give ChatGPT-4/CI a solid passing grade in introductory engineering math and electromagnetics.
- There is a major jump in performance from ChatGPT-3.5 to ChatGPT-4, and another major jump in performance from ChatGPT-4 to ChatGPT-4/CI. In particular, the use of Code Interpreter enabled accurate symbolic manipulations and numerical calculations.
- The performance of all versions of ChatGPT is somewhat stochastic. Within a single instance (a new "chat"), it is difficult to get ChatGPT to meaningfully check its work and correct its own mistakes without the user pointing out a specific mistake.
- Given the stochastic nature and "stickiness" described above, we found that the best way to get a correct answer on a particular problem is to solve the same problem N times (e.g., N = 10) in N separate ChatGPT-4/CI instances, and take the consensus (most-common) answer. Using this approach, all problems that we tested were solved correctly without any additional human input.
- While custom instructions ("prompt engineering") were useful in some cases where definitions of variables needed to be clarified or to make ChatGPT more verbose when explaining solutions, we did not identify any prompts that measurably improved the problem-solving performance of ChatGPT-4/CI on the problems we tested.

**Sections below**: 1. Short intro; 2. The problems; 3. Problem-solving performance; 4. Advice for students and instructors based on our findings; 5. Appendices; 6. References

[i] Notably, an image can be loaded and analyzed with a Python script written by ChatGPT and executed using Code Interpreter

## 1. Short intro

GPT-3[1] and GPT-4[2] are large language models (LLMs) released by OpenAI that have been shown to demonstrate at least some problem-solving abilities. These models been adapted for a conversational format in ChatGPT, which currently offers modes 3.5 and 4. Several plug-ins have also been introduced that can function as tools for ChatGPT; for example, the WolframAlpha plugin enables ChatGPT to send requests to WolframAlpha, which can then perform symbolic and arithmetic operations or retrieve certain facts.

We were specifically interested in the Code Interpreter plug-in (recently renamed Advanced Data Analysis), in which ChatGPT can write and execute Python code and use certain Python libraries. Code Interpreter compensates for many of ChatGPT's apparent weaknesses in quantitative problem-solving by writing and executing Python code, using libraries such as *math*, *NumPy*, *SymPy*, and others to perform symbolic manipulations and arithmetic operations. It can also generate plots, e.g., using *Matplotlib*. Notably, ChatGPT can self-correct compile-time errors, modifying the code to fix the error and re-running.

The results provided below built on a growing set of tests and benchmarks of GPT-3, GPT-4, and ChatGPT on various math and science problems[3,4], including recent testing using the WolframAlpha and Code Interpreter plugins[4]. Note also that there are now efforts to create datasets to systematically evaluate LLM performance on problem-solving (e.g., SciBench[5], SciEval[6], JEEBench[7]); our effort here is smaller in scale, but enabled us to look at each trial and identify failure modes and patterns.

## 2. The problems

The 13 problems were selected somewhat arbitrarily from typical problems that UW-Madison ECE students in introductory electromagnetics were expected to complete during in-class flipped-learning sessions or on a quiz. The provenance of many of the problems is not fully known, but it is likely that some of the problems were inspired by textbook problems (in particular, Ulaby's Fundamentals of Applied Electromagnetics) with some modifications, while others are original. However, even original problems are likely similar to those that can be found in various textbooks, since there is a limited set of problems in basic electromagnetics that can be solved by students in a portion of a class period.

We emphasize that these problems were selected before they were evaluated with ChatGPT. That is, we did not remove any problems after observing how ChatGPT performed on them, with the exception of two problems which turned out to be more ambiguous than we expected, and therefore were more-challenging to grade (qualitatively the performance of ChatGPT on those two problems was similar to its performance on the problems we present here).

The problems are as follows, formatted for ease of human readability; the exact prompts fed into ChatGPT are given in Appendix 1.

1. Given matrices $\boldsymbol{A}$ and $\boldsymbol{B}$, find $(\boldsymbol{A} - \boldsymbol{B})$

   $\boldsymbol{A} = \begin{bmatrix} -5 & -5 & -4 \\ -1 & -3 & -4 \end{bmatrix}$

   $\boldsymbol{B} = \begin{bmatrix} -9 & -3 & -10 \\ -2 & -7 & -2 \end{bmatrix}$

2. Given $\boldsymbol{A} = \begin{bmatrix} 6.5 & 6.6 & 3.9 & 3.8 \\ 0 & 6 & 7.9 & 3.1 \\ 0 & 0 & 3.3 & 5.3 \\ 0 & 0 & 0 & 3 \end{bmatrix}$, find $|\boldsymbol{A}|$

3. Calculate the total charge in a cube, 1 meter long on each side, located in the first octant ($x \geq 0, y \geq 0, z \geq 0$), with one corner at the origin. The volume charge density in the cube is $\rho_v = xy^2 e^{-3z}$ mC/m$^3$. Start with writing down the charge density neatly first.

4. A line of electric charge with uniform density $\rho = 3$ μC/m exists along the z-axis between $z = 0$ and $z = 9$ meters. Find $\boldsymbol{E}$ in vector form at $R(x, y, z) = (0, 22, 0)$ meters.

5. Calculate the divergence (strength of flux sources) of the vector field $\boldsymbol{A} = 2\sqrt{x^2 + y^2}\,\hat{x}$ at the point $(x, y, z) = (2, 6, -7)$. Give three significant figures.

6. Calculate the divergence (strength of the flux sources) of the vector field $\boldsymbol{A} = r sin(2\phi)\,\hat{r} + r cos(2\phi)\,\hat{\phi} + (z^2)\hat{z}$ at the point $(r, \phi, z) = \left(19, \frac{2\pi}{18}, 1\right)$. Give three significant figures.

7. You have potential $V = Ar^2 sin(\phi) + Brz$, where $A$ and $B$ are constants with the appropriate units. Calculate the electric flux density $\boldsymbol{D}$ corresponding to this potential.

8. A thin hemispheric shell in the upper half-plane ($R = a, 0 \leq \theta \leq \frac{\pi}{2}, 0 \leq \phi \leq 2\pi$) has uniform surface charge density $\rho_s$. Find the potential at the origin, assuming the potential is 0 infinitely far away.

9. Two small identical spheres have charges of $+20$ nC and $-5$ nC, respectively.

   a. What is the force between them if they are separated by 10 cm?

   b. The two spheres are brought into contact with each other (such that they equalize their charge due to conduction between the spheres) and then separated again by 10 cm. What is the force between them now?

10. You have a sphere of radius $a$, inside which the charge density can be written as $\rho_v = \frac{\rho_0}{R^2}$. Outside, the charge is zero. Find the electric field $\boldsymbol{E}$ outside the sphere.

11. The plane defined by $y = 0$ separates two dielectrics. In $y < 0$, we have medium 1, with permittivity $\epsilon_0$. In $y > 0$, we have medium 2, with permittivity $\epsilon = \epsilon_r \epsilon_0$. There is no charge on the interface. Just next to the interface, the field in medium 1 is $\boldsymbol{E_1} = a\hat{x} + b\hat{y}$. Find $\boldsymbol{E_2}$ in medium 2, just next to the interface.

12. A coaxial cable is made of two thin shells, each made of a perfect electric conductor. The inner shell radius is $r_1$, and the outer shell radius is $r_2$. In between the shells is air (you can assume it's free space). The length of the cable is $L \gg r_2$.

a. (10 pts) Put charge $+Q$ on the inner plate, and $-Q$ on the outer plate. Calculate field everywhere in space (not just in between the plates!).

b. (5 pts) Find the capacitance, $C$, of this configuration, based on the result in (a). Make sure to show your work step by step!

13. You have a rectangular conducting loop in three-dimensional space. One side of the loop is along the z-axis with one end on the origin, with length $b = 3$ cm. Another side of the loop is in the x-y plane and has length $a = 2$ cm. The loop is a rectangle. This loop is rotating around the z-axis at 6,000 RPM, and has internal resistance $R = 0.5\ \Omega$. The orientation of the loop can be defined with an angle $\phi$ from the x axis. There is an external magnetic field which is constant, and pointing in the y-direction everywhere, so $B = B_0\hat{y}$, where $B_0 = 1$ T.

a. What is the angular frequency, $\omega$, and what is the expression for $\phi(t)$ assuming that $\phi(t = 0) = 0$?

b. Find the expression for the magnetic flux, $\phi(t)$, through the loop, in terms of $a$, $b$, $B_0$, $\omega$, and time $t$.

c. Find the expression for the current flowing through the loop. Plot the current flowing through the loop vs. time, $I(t)$. On your plot, actually label the maximum and minimum current by plugging in all of the numbers.

## 3. Problem-solving performance

In the table below, we summarize the performance of the different modes of ChatGPT on the 13 problems given in the previous section. Initially, we were not sure how many trials we would need, so we experimented. By the end, we realized that 10 trials were sufficient to get an understanding of ChatGPT's performance. For this reason, most of the problems were attempted 10 times for each mode of ChatGPT, but several initial ones (problems 1 and 2, for some modes, and problem 13) were attempted more times than that.

Note that we also tested a few custom prompts on some of the problems, but we are not including these tests in the main text, because we found little difference in performance with and without all of the custom prompts we tried; the full results can be found in Appendix 2.

| **Problem** | **ChatGPT-3.5**[a] | **ChatGPT-4** | **ChatGPT-4/CI** |
|---|---|---|---|
| 1. Matrix subtraction | 9/17 | 9/10 | 10/10 |
| 2. Upper triangular matrix determinant | 4/15 | 9/15 | 10/10 |
| 3. Integrating charge density in Cartesian coordinates | 2/10[b] | 10/10 | 10/10 |
| 4. Electric field from line charge in Cartesian coordinates | 0/10 | 0/10[c] | 6/10 |
| 5. Divergence in Cartesian coordinates | 5/10 | 8/10 | 9/10 |
| 6. Divergence in cylindrical coordinates | 5/10 | 7/10 | 7/10 |
| 7. Electric flux density ($D$) from potential (V) in cylindrical coordinates | 1/10 | 3/10 | 7/10 |
| 8. Potential of hemispherical shell at geometric center | 0/10 | 4/10 | 7/10 |
| 9. Force between two charges | 0/10 | 1/10 | 10/10 |
| 10. Electric field outside charged sphere | 0/10 | 8/10 | 8/10 |
| 11. Electric field at dielectric interface | 0/10 | 10/10 | 10/10 |
| 12. Electric field and capacitance of coaxial cable | 3/10 | 7/10[d] | 7/10 |
| 13. Induced current in rotating rectangular conducting loop in a magnetic field | 1/10 | 0/10 | 16/30[e] |

[a] Cases where ChatGPT-3.5 stopped at the correct analytical form of the solution with instructions on how to plug in numbers and proceed were marked as correct.

[b] ChatGPT-3.5 does well on problem 3 in forming the integrals; it does poorly in solving the integrals.

[c] ChatGPT-4 does well on problem 4 in forming the integral; it does poorly in solving the integral.

[d] Includes negative capacitance marked as correct.

[e] Problem 13 with ChatGPT-4/CI has 30 trials because we faced network issues during the first 10 trials, resulting in text generation stopping before reaching final answer. To check if these issues were affecting our data, we ran 20 more trials. Of these 30 trials, 16 were unambiguously correct. Of the 14 marked wrong, There were 6 cases where ChatGPT-4/CI worked out the problem correctly (including the plot), but wrongly stated the final answer after the plot. See Fig. 12 for example.

Note that the most-common (consensus) result GPT-4/CI is correct on every problem we tested. This means that a user can solve each problem e.g. 10 times in 10 different instances using GPT-4/CI and take the consensus answers, yielding the correct answer in each of the 13 problems we tested.

## 4. Advice for students and instructors based on our findings

Advice for instructors:

- Be aware that the capabilities of generative AI with plug-ins are rapidly increasing. Instructors are advised to test out leading AI tools at least once a semester to stay on top of their capabilities. As a specific example, the introduction of Code Interpreter in July 2023 significantly improved the problem-solving capacity of ChatGPT, so user experience with earlier versions does not generalize.
- It is recommended that instructors test problems for completion at home with the latest versions of generative AI tools (and with the relevant plugins) to gauge the degree to which AI tools can complete large portions of the assignments or even entire assignments.
- Assessment of unaided student performance (i.e., without AI tools) should be done in person as much as possible, for example with in-class examinations.
- Instructors may consider sharing their findings of the capabilities of AI tools on the topics in the course with the students (and other instructors). This helps avoid equity issues, where some fraction of the students use AI tools on assignments and to study, but others are not aware of the tools' evolving capabilities.
- Instructors may explore generative AI tools to generate practice problems, help write well-formatted solutions, and otherwise enhance the learning experience.
- Instructors may consider creating at least some new assignments and projects that allow the use of AI tools, to enhance student understanding of the tools that will be available to them in the future, and to enable more-ambitious projects.

Advice for students:

- It is to your advantage to learn how to solve problems both (a) without AI tools, and (b) with AI tools. An imperfect analogy is the use of calculators, which did not remove the need for students to learn how to add, subtract, and multiply in elementary school. The ability to solve problems without AI tools is part of the necessary foundation for higher-level thinking and for the engineering of new technologies.
- Meanwhile, there may be value in using the most-recent AI tools to enhance your learning. Tools like ChatGPT, if properly benchmarked on the content of a course, can be used as a tutor and aid. And ability to solve problems with AI tools is a skill that will undoubtedly become more important in the near future.
- Tools like ChatGPT can provide solutions to problems that look believable, but sometimes contain errors. Make sure to carefully benchmark any tool you use against your own understanding and previously solved problems to ensure that the tool is giving you correct solutions rather than solutions that only appear to be correct. As with any other tool, seek to understand its capabilities and limitations.
- Because tools like ChatGPT can be confidently incorrect, and will sometimes even defend such incorrect answers, it can be a good exercise to carefully compare ChatGPT solutions with your own and figure out which solution is correct and why.

### 5.1. Appendix: Sample successes and failures:

Here we present a small subset of the ChatGPT outputs we generated, showing both successful solutions and failures. Because there is currently not a good way to export ChatGPT outputs using Code Interpreter (now renamed Advanced Data Analysis), we provide screenshots of the output in two-column format. Everywhere where there is a mistake, we identify the first place the mistake is made with a red circle or box. Green boxes or checkmarks indicate places where the answer is correct, whether it is before a mistake is made or after a correction to a mistake is made.

Example of a success with ChatGPT-4/CI, on Problem 13, demonstrating analytical and plotting abilities:

TA

You have a rectangular conducting loop in three-dimensional space. One side of the loop is along the z axis with one end on the origin, with length b = 3 cm. Another side of the loop is in the x-y plane and has length a = 2 cm. The loop is a rectangle.
This loop is rotating around the z axis at 6,000 RPM, and has internal resistance R=0.5 Ω.
The orientation of the loop can be defined with an angle ϕ from the x axis. There is an external magnetic field which is constant, and pointing in the y direction everywhere, so B=B_0 y^, where B_0=1 T.
a. What is the angular frequency, ω, and what is the expression for ϕ(t) assuming that ϕ(t=0)=0?
b. Find the expression for the magnetic flux, Φ(t), through the loop, in terms of a,b,B_0,ω, and time t.
c. Find the expression for the current flowing through the loop. Plot the current flowing through the loop vs. time, I(t). On your plot, actually label the maximum and minimum current by plugging in all of the numbers.

Alright, let's break this down step by step:

**a. Angular frequency and $\phi(t)$**

The angular frequency $\omega$ is given by:

$\omega = 2\pi f$

where $f$ is the frequency in revolutions per second.

Given that the rotation is 6000 RPM (revolutions per minute), we can convert this to revolutions per second:

$f = \frac{6000}{60}$

Now, we can find $\omega$.

Once we have $\omega$, the expression for $\phi(t)$ is given by:

$\phi(t) = \omega t$

since $\phi(t = 0) = 0$.

**b. Magnetic flux $\Phi(t)$**

The magnetic flux $\Phi$ through the loop at any given time is the product of the magnetic field $B_0$, the area $A$ of the loop, and the cosine of the angle $\phi(t)$ between the normal to the loop's plane and the magnetic field. In this case, since the magnetic field is along the y-direction:

$$\Phi(t) = B_0 \times A \times \cos(\phi(t))$$

Where:

$A = a \times b$

**c. Current through the loop**

The current $I$ flowing through the loop due to a change in magnetic flux is given by Faraday's law:

$\varepsilon = -\frac{d\Phi(t)}{dt}$

where $\varepsilon$ is the induced electromotive force (EMF).

The current $I$ is then:

$I(t) = \frac{\varepsilon}{R}$

where $R$ is the resistance of the loop.

Let's solve for each of these in order:

Finished working Hide work

python Copy code

```python
import math

# Given values
b = 0.03  # in meters
a = 0.02  # in meters
RPM = 6000  # revolutions per minute

# a. Angular frequency and φ(t)
f = RPM / 60  # converting RPM to revolutions per second
omega = 2 * math.pi * f  # angular frequency in radians per second

omega
```

RESULT

```
628.3185307179587
```

a. The angular frequency $\omega$ is approximately $628.32$ radians/second.

Now, for the expression $\phi(t)$, it's simply:

$\phi(t) = \omega t$

Next, we'll find the magnetic flux $\Phi(t)$.

**b. Magnetic flux $\Phi(t)$**

The area $A$ of the loop is $a \times b$, and the magnetic flux through the loop is given by:

$\Phi(t) = B_0 \times A \times \cos(\phi(t))$

Let's calculate $A$ and then the expression for $\Phi(t)$.

Finished working Hide work

*Continued on next page*

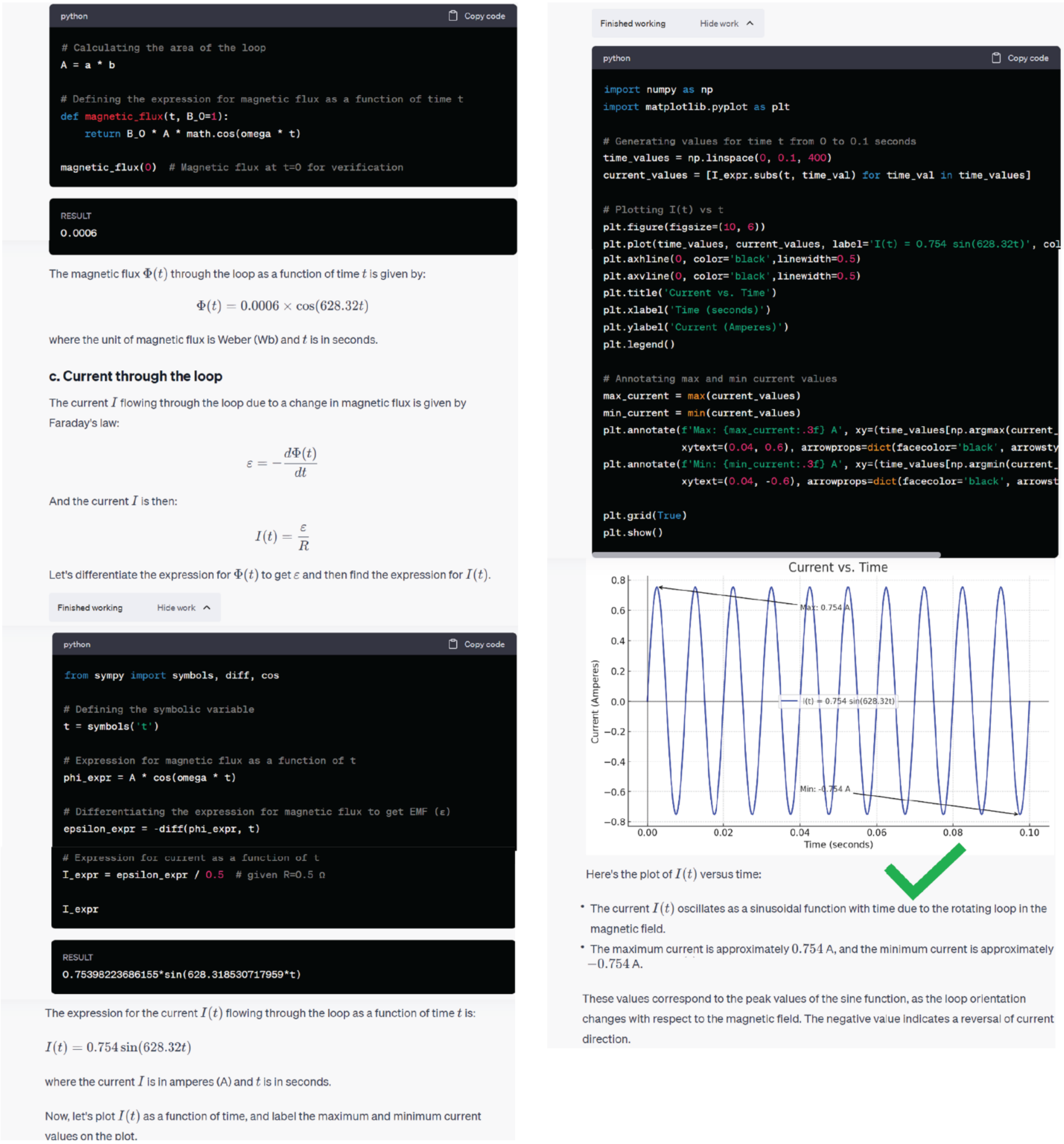

Fig. 1: An example of ChatGPT-4/CI working its way through Problem 13, a problem that also requires creating a plot. Note the use of Code Interpreter for mathematical operations as well as plotting. While ChatGPT-4/CI is good at plotting, it lacks spatial awareness, and therefore the labels are in awkward positions.

Example of a success with ChatGPT-4/CI, on Problem 4, demonstrating analytical abilities:

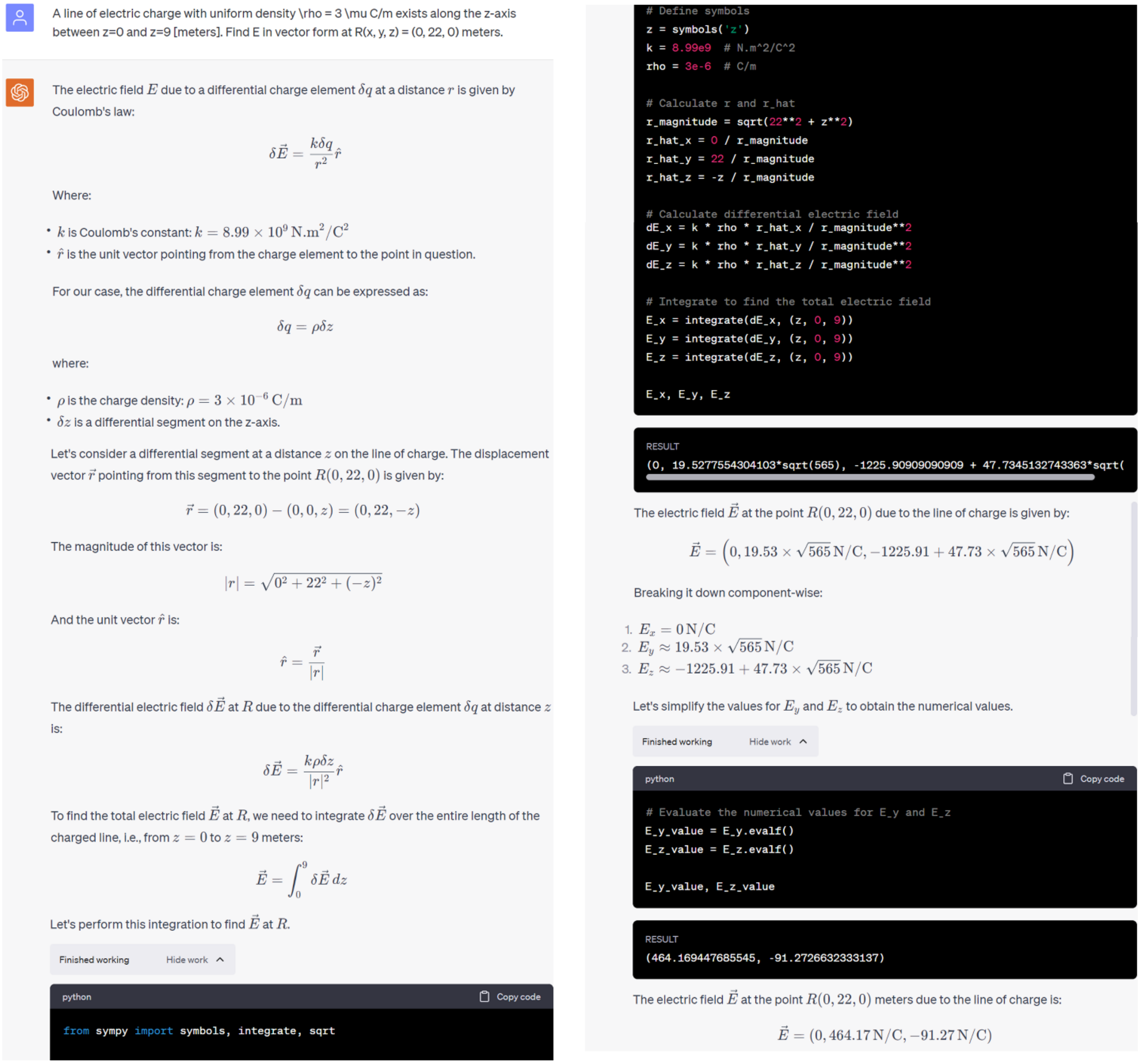

A line of electric charge with uniform density \rho = 3 \mu C/m exists along the z-axis between z=0 and z=9 [meters]. Find E in vector form at R(x, y, z) = (0, 22, 0) meters.

The electric field $E$ due to a differential charge element $\delta q$ at a distance $r$ is given by Coulomb's law:

$$\delta \vec{E} = \frac{k \delta q}{r^2} \hat{r}$$

Where:

- $k$ is Coulomb's constant: $k = 8.99 \times 10^9 \, \mathrm{N.m^2/C^2}$
- $\hat{r}$ is the unit vector pointing from the charge element to the point in question.

For our case, the differential charge element $\delta q$ can be expressed as:

$$\delta q = \rho \delta z$$

where:

- $\rho$ is the charge density: $\rho = 3 \times 10^{-6} \, \mathrm{C/m}$
- $\delta z$ is a differential segment on the z-axis.

Let's consider a differential segment at a distance $z$ on the line of charge. The displacement vector $\vec{r}$ pointing from this segment to the point $R(0, 22, 0)$ is given by:

$$\vec{r} = (0, 22, 0) - (0, 0, z) = (0, 22, -z)$$

The magnitude of this vector is:

$$|r| = \sqrt{0^2 + 22^2 + (-z)^2}$$

And the unit vector $\hat{r}$ is:

$$\hat{r} = \frac{\vec{r}}{|r|}$$

The differential electric field $\delta \vec{E}$ at $R$ due to the differential charge element $\delta q$ at distance $z$ is:

$$\delta \vec{E} = \frac{k \rho \delta z}{|r|^2} \hat{r}$$

To find the total electric field $\vec{E}$ at $R$, we need to integrate $\delta \vec{E}$ over the entire length of the charged line, i.e., from $z = 0$ to $z = 9$ meters:

$$\vec{E} = \int_0^9 \delta \vec{E} \, dz$$

Let's perform this integration to find $\vec{E}$ at $R$.

Finished working Hide work

python Copy code

```python
from sympy import symbols, integrate, sqrt

# Define symbols
z = symbols('z')
k = 8.99e9  # N.m^2/C^2
rho = 3e-6  # C/m

# Calculate r and r_hat
r_magnitude = sqrt(22**2 + z**2)
r_hat_x = 0 / r_magnitude
r_hat_y = 22 / r_magnitude
r_hat_z = -z / r_magnitude

# Calculate differential electric field
dE_x = k * rho * r_hat_x / r_magnitude**2
dE_y = k * rho * r_hat_y / r_magnitude**2
dE_z = k * rho * r_hat_z / r_magnitude**2

# Integrate to find the total electric field
E_x = integrate(dE_x, (z, 0, 9))
E_y = integrate(dE_y, (z, 0, 9))
E_z = integrate(dE_z, (z, 0, 9))

E_x, E_y, E_z
```

RESULT

```
(0, 19.5277554304103*sqrt(565), -1225.90909090909 + 47.7345132743363*sqrt(
```

The electric field $\vec{E}$ at the point $R(0, 22, 0)$ due to the line of charge is given by:

$$\vec{E} = \left(0, 19.53 \times \sqrt{565} \, \mathrm{N/C}, -1225.91 + 47.73 \times \sqrt{565} \, \mathrm{N/C}\right)$$

Breaking it down component-wise:

1. $E_x = 0 \, \mathrm{N/C}$
2. $E_y \approx 19.53 \times \sqrt{565} \, \mathrm{N/C}$
3. $E_z \approx -1225.91 + 47.73 \times \sqrt{565} \, \mathrm{N/C}$

Let's simplify the values for $E_y$ and $E_z$ to obtain the numerical values.

Finished working Hide work

python Copy code

```python
# Evaluate the numerical values for E_y and E_z
E_y_value = E_y.evalf()
E_z_value = E_z.evalf()

E_y_value, E_z_value
```

RESULT

```
(464.169447685545, -91.2726632333137)
```

The electric field $\vec{E}$ at the point $R(0, 22, 0)$ meters due to the line of charge is:

$$\vec{E} = (0, 464.17 \, \mathrm{N/C}, -91.27 \, \mathrm{N/C})$$

Fig. 2: Example of ChatGPT-4/CI successfully working its way through Problem 4, a problem that other versions of ChatGPT were unable to correctly solve.

Example of the stochastic nature of ChatGPT: two instances of the same problem give different answers:

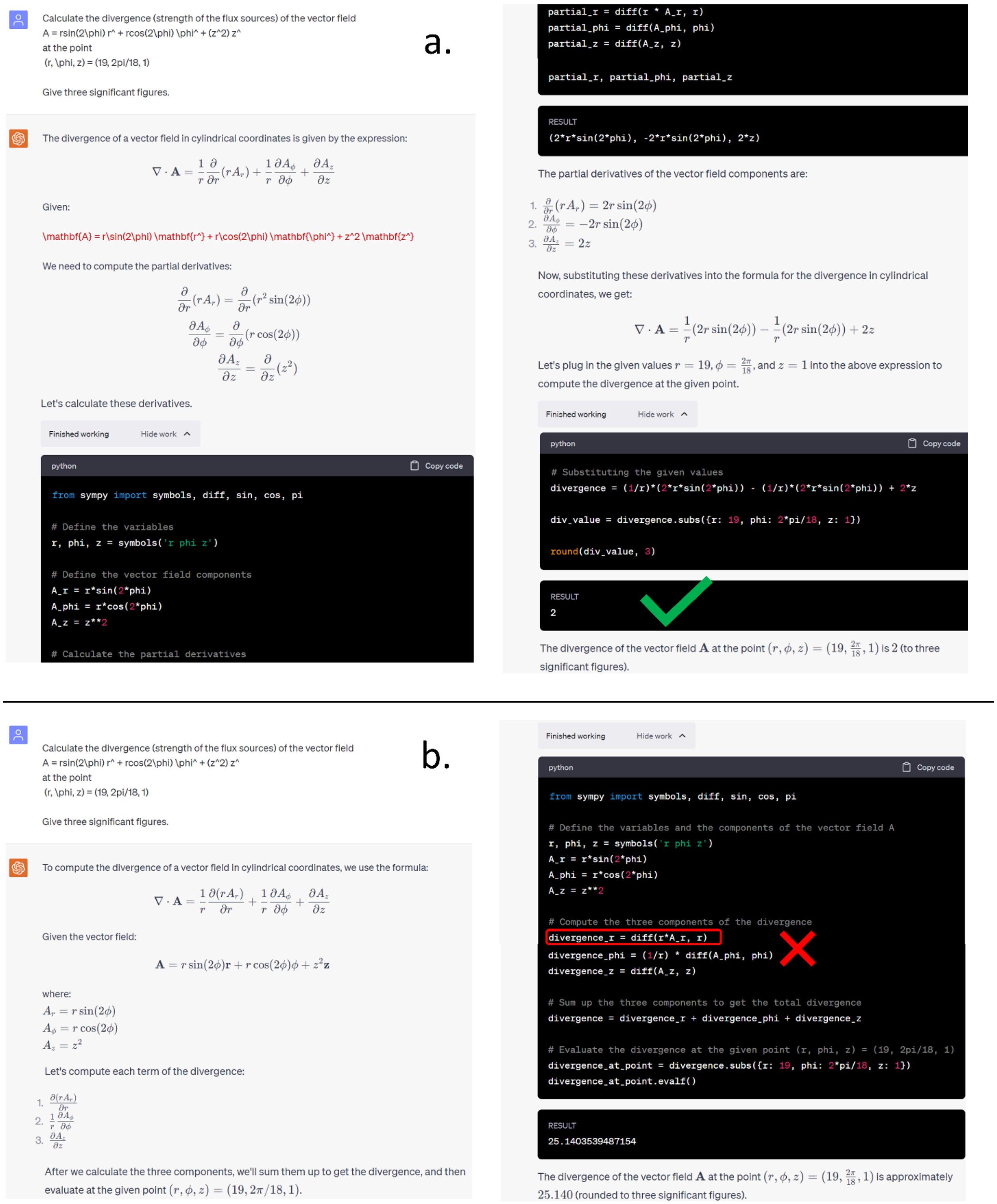

Fig. 3: A comparison of two instances (chats) of ChatGPT-4/CI giving two different answers ((a) correct, (b) incorrect) for the same question. This stochastic nature of ChatGPT was observed for all versions, except for simple problems (1, 2, 3, 9 and 11) where ChatGPT-4/CI scored 10/10 each.

Code Interpreter makes up for the arithmetic abilities lacking in ChatGPT-4:

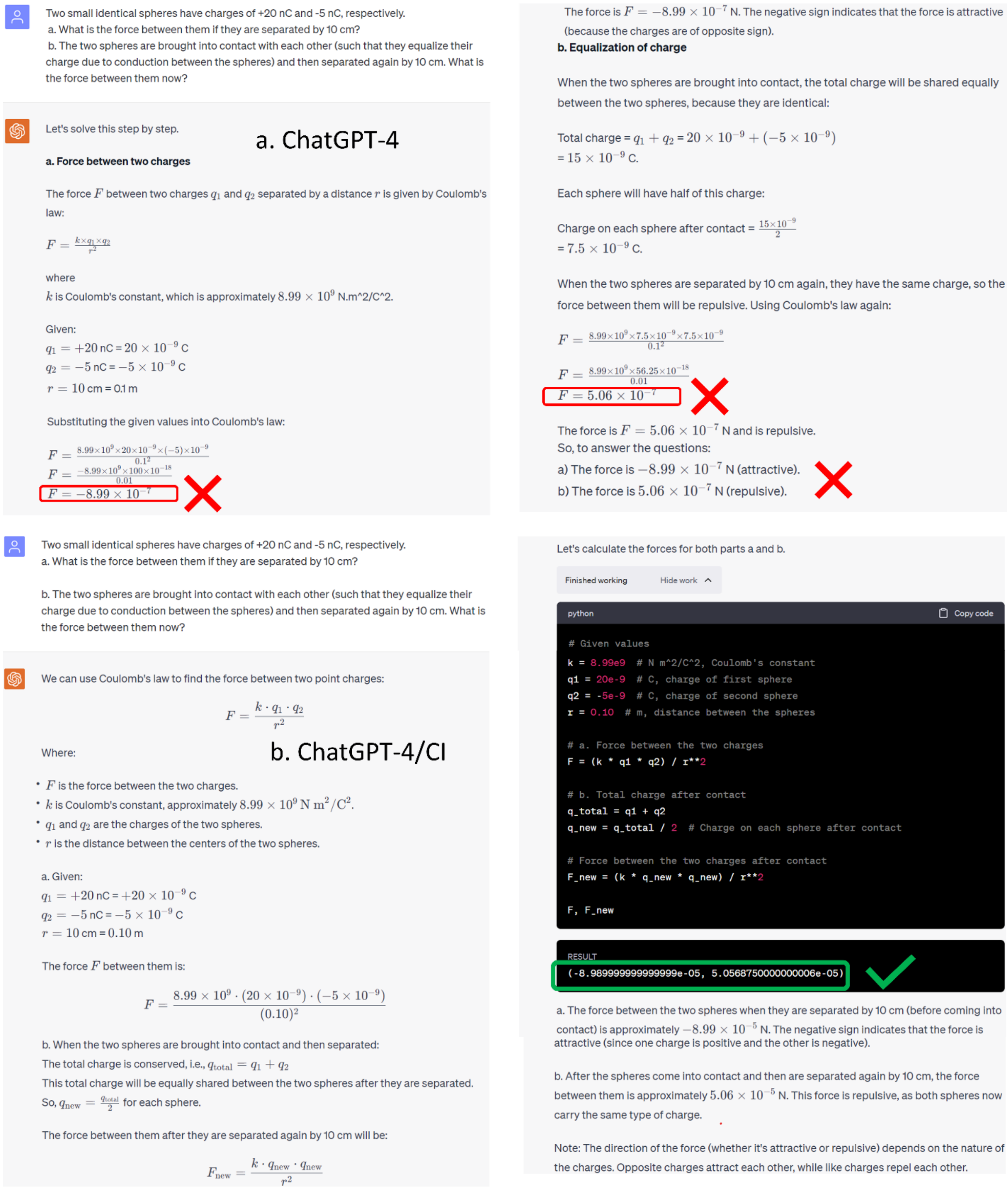

Fig. 4: In our experiments, ChatGPT-4/CI gave correct answers more consistently than ChatGPT-4 because of the code interpreter. In this example using Problem 9, ChatGPT-4 (a) was able to analytically get the correct expression but gave the wrong order of magnitude in the numerical answer, whereas ChatGPT-4/CI (a) gave the correct numerical value using Code Interpreter.

Failure: ChatGPT-3.5 sometimes changes signs of integers:

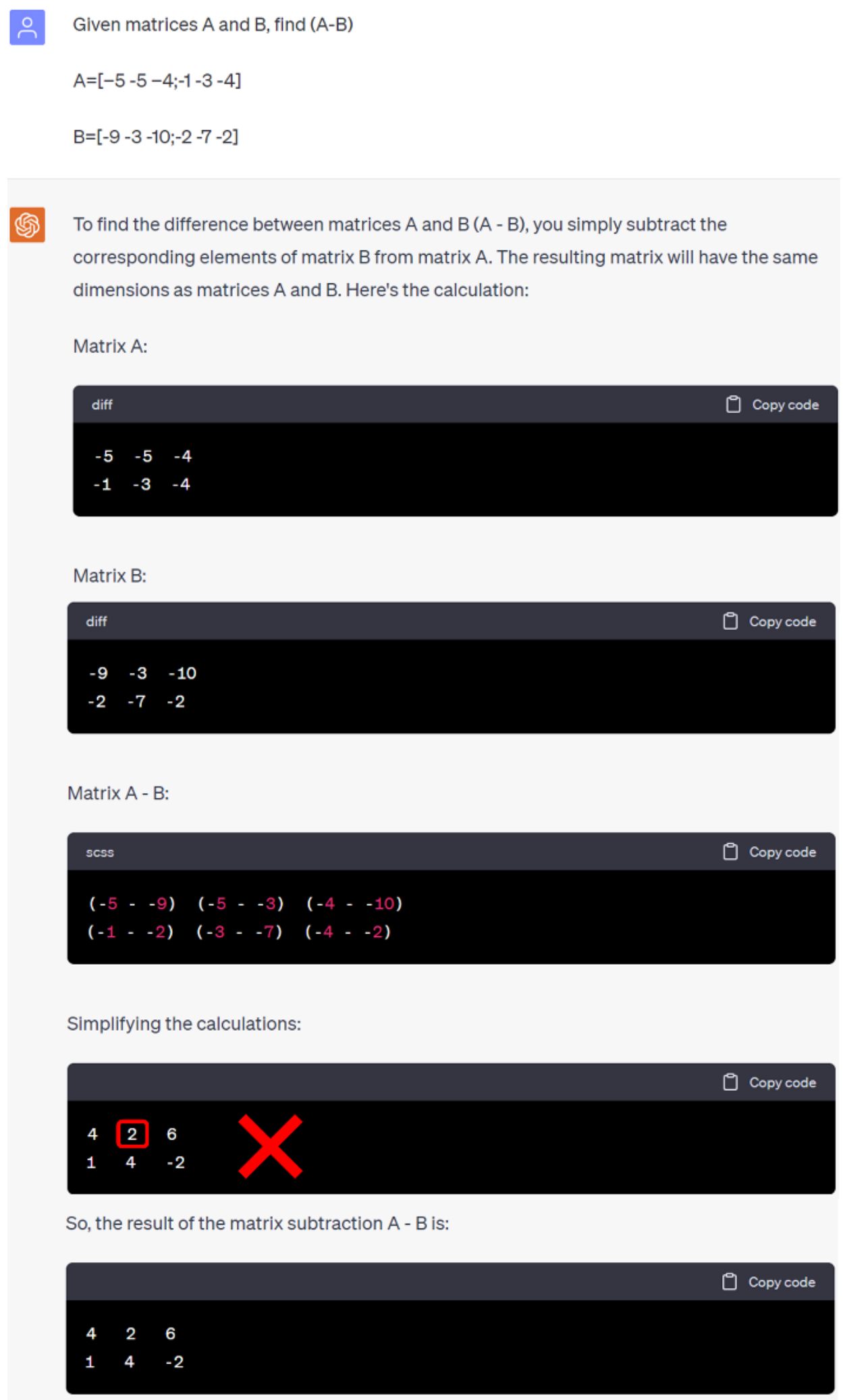

Fig. 5: An example of ChatGPT-3.5 making a mistake by changing the sign of an integer during simple matrix subtraction when solving Problem 1 (matrix subtraction). We observed that when this happened, the absolute value of the integer was usually correct.

Failure: ChatGPT-4 sometimes makes mistakes in multiplying more than 2 numbers simultaneously:

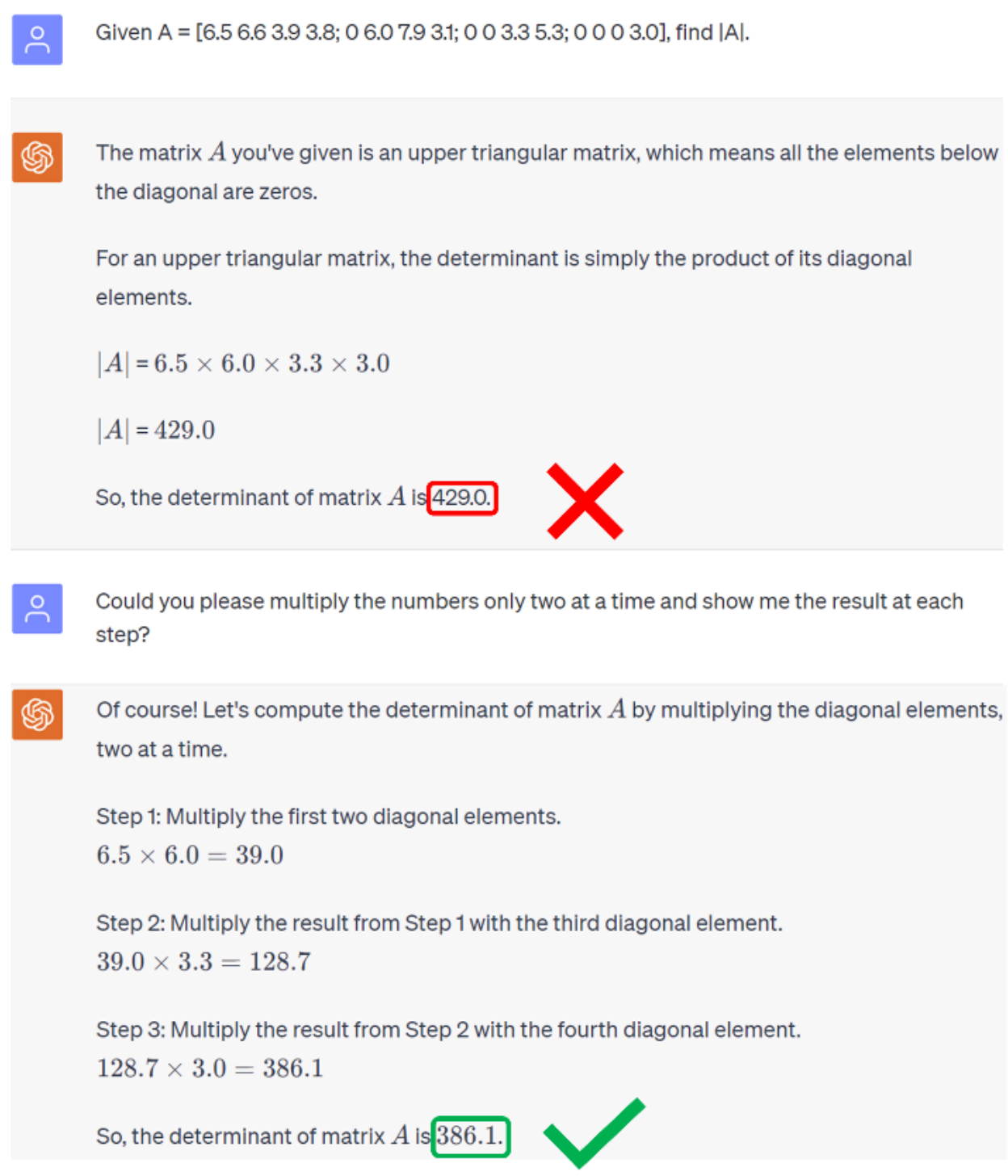

Fig. 6: An example where ChatGPT-4 wrongly multiplies 4 numbers simultaneously (Problem 2). Asking ChatGPT-4 to only multiply two numbers at a time (as shown in the figure) gave correct answers more often, though not always. We observed a similar behavior with ChatGPT-3.5

Failure: ChatGPT-3.5 can set up Cartesian integrals fairly well, but often cannot evaluate them:

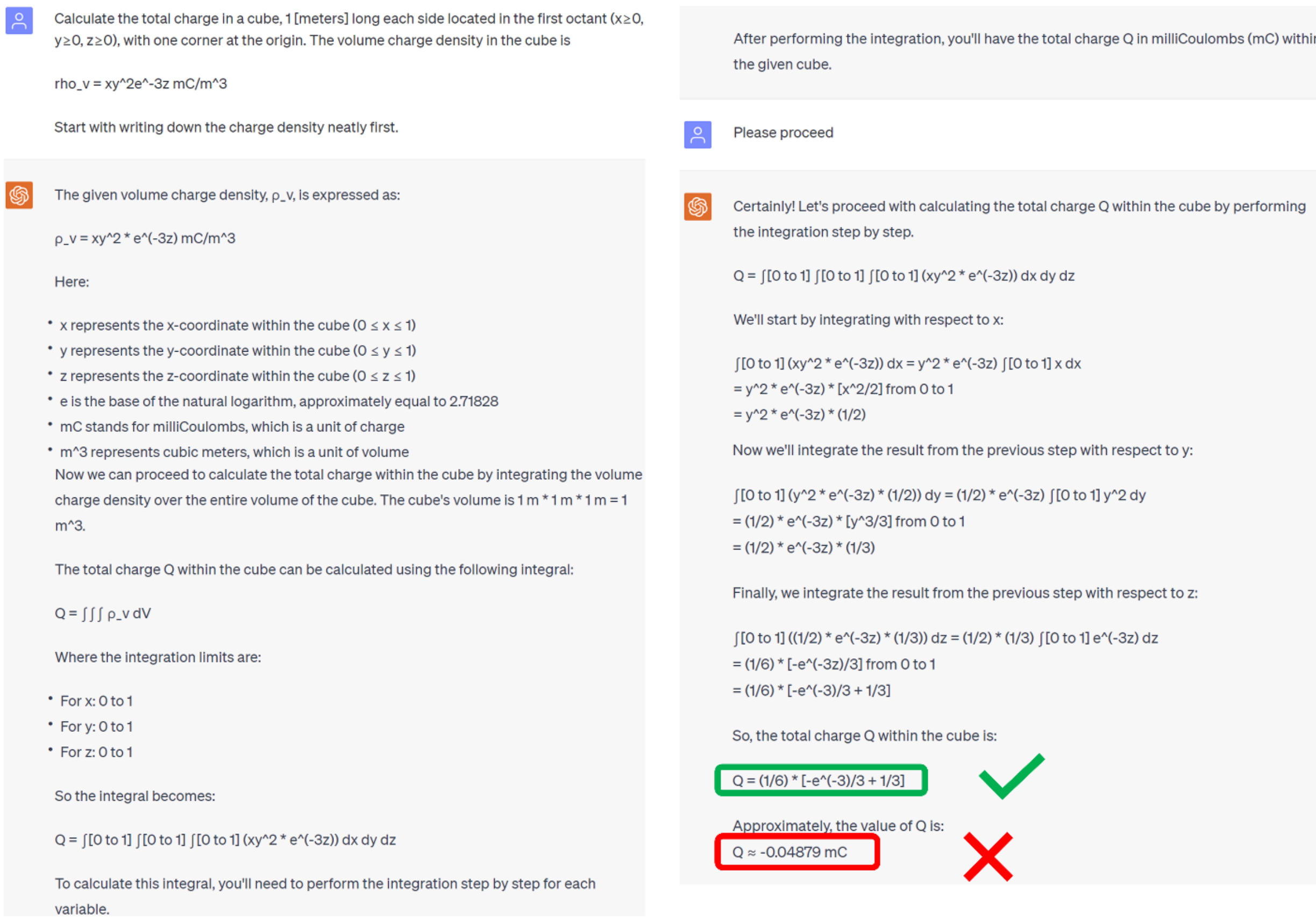
Calculate the total charge in a cube, 1 [meters] long each side located in the first octant (x≥0, y≥0, z≥0), with one corner at the origin. The volume charge density in the cube is

rho_v = xy^2e^-3z mC/m^3

Start with writing down the charge density neatly first.

The given volume charge density, ρ_v, is expressed as:

ρ_v = xy^2 * e^(-3z) mC/m^3

Here:

- x represents the x-coordinate within the cube (0 ≤ x ≤ 1)
- y represents the y-coordinate within the cube (0 ≤ y ≤ 1)
- z represents the z-coordinate within the cube (0 ≤ z ≤ 1)
- e is the base of the natural logarithm, approximately equal to 2.71828
- mC stands for milliCoulombs, which is a unit of charge
- m^3 represents cubic meters, which is a unit of volume

Now we can proceed to calculate the total charge within the cube by integrating the volume charge density over the entire volume of the cube. The cube's volume is 1 m * 1 m * 1 m = 1 m^3.

The total charge Q within the cube can be calculated using the following integral:

Q = ∫∫∫ ρ_v dV

Where the integration limits are:

- For x: 0 to 1
- For y: 0 to 1
- For z: 0 to 1

So the integral becomes:

Q = ∫[0 to 1] ∫[0 to 1] ∫[0 to 1] (xy^2 * e^(-3z)) dx dy dz

To calculate this integral, you'll need to perform the integration step by step for each variable.

After performing the integration, you'll have the total charge Q in milliCoulombs (mC) within the given cube.

Please proceed

Certainly! Let's proceed with calculating the total charge Q within the cube by performing the integration step by step.

Q = ∫[0 to 1] ∫[0 to 1] ∫[0 to 1] (xy^2 * e^(-3z)) dx dy dz

We'll start by integrating with respect to x:

∫[0 to 1] (xy^2 * e^(-3z)) dx = y^2 * e^(-3z) ∫[0 to 1] x dx
= y^2 * e^(-3z) * [x^2/2] from 0 to 1
= y^2 * e^(-3z) * (1/2)

Now we'll integrate the result from the previous step with respect to y:

∫[0 to 1] (y^2 * e^(-3z) * (1/2)) dy = (1/2) * e^(-3z) ∫[0 to 1] y^2 dy
= (1/2) * e^(-3z) * [y^3/3] from 0 to 1
= (1/2) * e^(-3z) * (1/3)

Finally, we integrate the result from the previous step with respect to z:

∫[0 to 1] ((1/2) * e^(-3z) * (1/3)) dz = (1/2) * (1/3) ∫[0 to 1] e^(-3z) dz
= (1/6) * [-e^(-3z)/3] from 0 to 1
= (1/6) * [-e^(-3)/3 + 1/3]

So, the total charge Q within the cube is:

Q = (1/6) * [-e^(-3)/3 + 1/3]

Approximately, the value of Q is:
Q ≈ -0.04879 mC

Fig. 7: An example of how ChatGPT-3.5 can often set up equations, but is then unable to correctly evaluate them. This example is on Problem 3 (calculating total charge from a charge distribution in cartesian coordinates).

Failure: ChatGPT-4/CI will occasionally make mistakes like dropping a term in correct solutions that were found with Code Interpreter:

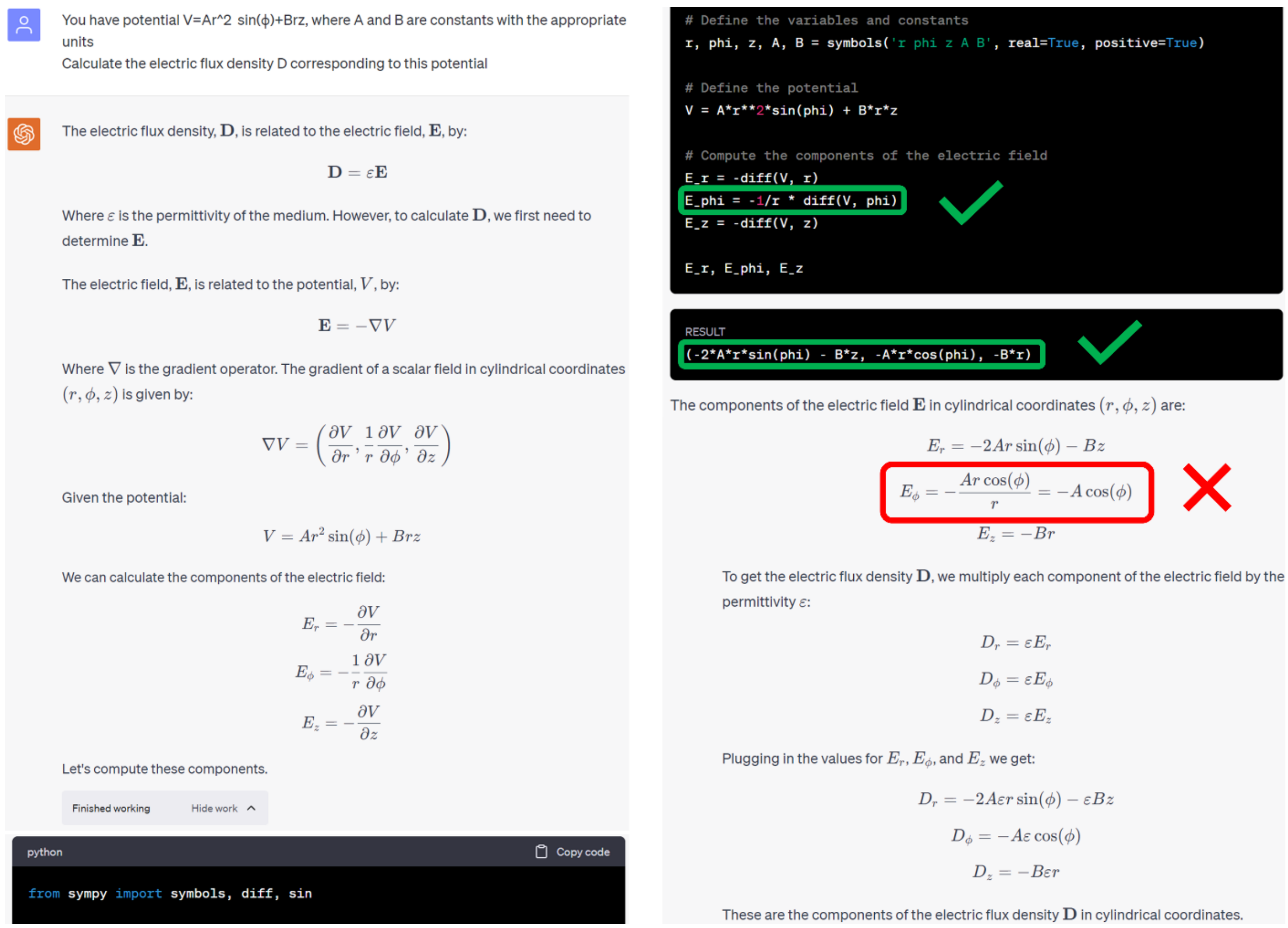

Fig. 8: Screenshot showing how ChatGPT-4/CI, despite having access to the code-interpreter, can make simple mistakes such as dividing the $E_\phi$ term by $r$ twice (once in the code, once just after the code). This example uses Problem 7, calculating the displacement field from a potential in cylindrical coordinates.

Failure: ChatGPT-4/CI has problems figuring out the distance from center of hemisphere to surface:

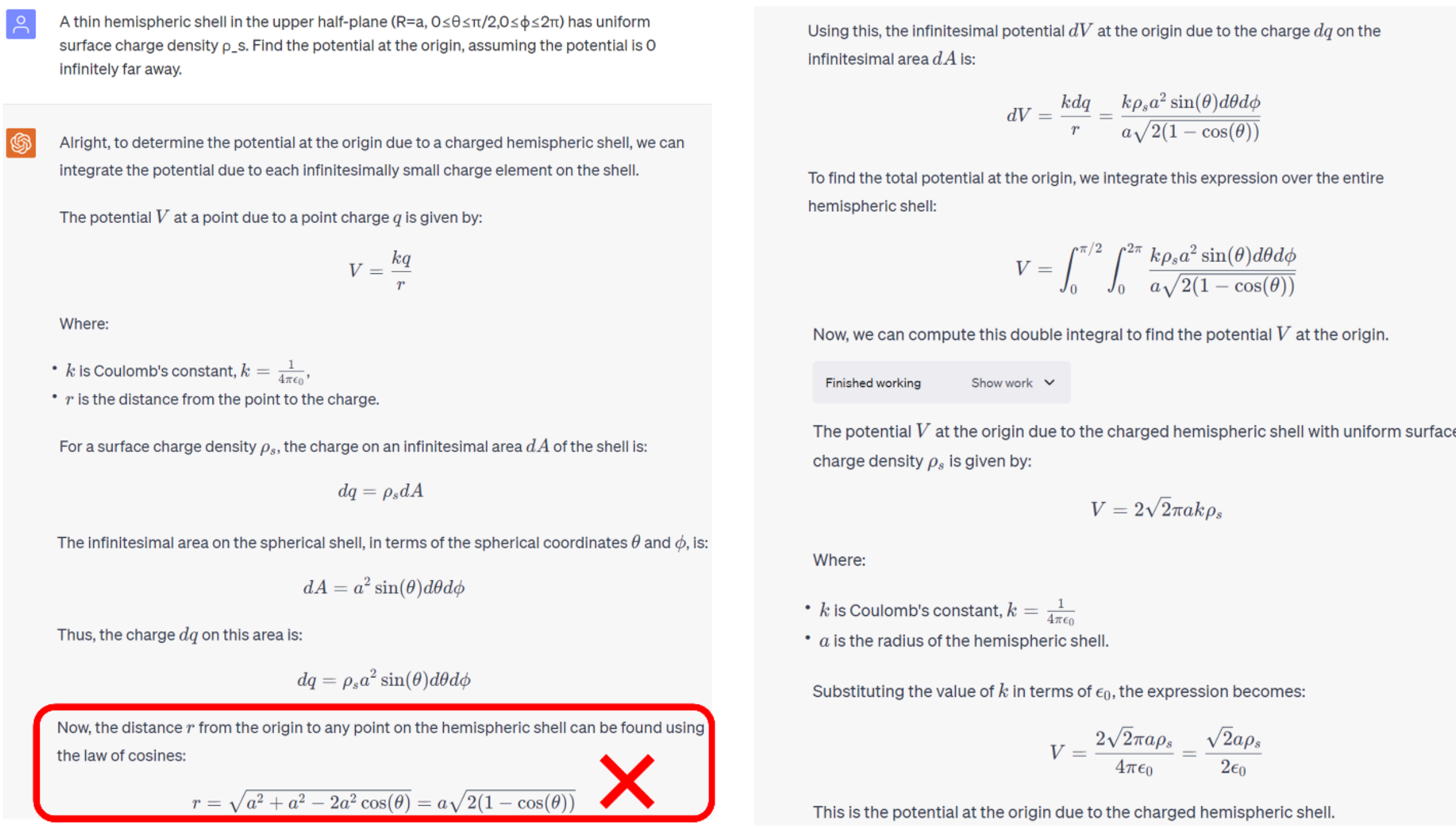

A thin hemispheric shell in the upper half-plane (R=a, 0≤θ≤π/2,0≤ϕ≤2π) has uniform surface charge density ρ_s. Find the potential at the origin, assuming the potential is 0 infinitely far away.

Alright, to determine the potential at the origin due to a charged hemispheric shell, we can integrate the potential due to each infinitesimally small charge element on the shell.

The potential $V$ at a point due to a point charge $q$ is given by:

$$V = \frac{kq}{r}$$

Where:

- $k$ is Coulomb's constant, $k = \frac{1}{4\pi\epsilon_0}$,
- $r$ is the distance from the point to the charge.

For a surface charge density $\rho_s$, the charge on an infinitesimal area $dA$ of the shell is:

$$dq = \rho_s dA$$

The infinitesimal area on the spherical shell, in terms of the spherical coordinates $\theta$ and $\phi$, is:

$$dA = a^2 \sin(\theta) d\theta d\phi$$

Thus, the charge $dq$ on this area is:

$$dq = \rho_s a^2 \sin(\theta) d\theta d\phi$$

Now, the distance $r$ from the origin to any point on the hemispheric shell can be found using the law of cosines:

$$r = \sqrt{a^2 + a^2 - 2a^2\cos(\theta)} = a\sqrt{2(1-\cos(\theta))}$$

Using this, the infinitesimal potential $dV$ at the origin due to the charge $dq$ on the infinitesimal area $dA$ is:

$$dV = \frac{kdq}{r} = \frac{k\rho_s a^2 \sin(\theta) d\theta d\phi}{a\sqrt{2(1-\cos(\theta))}}$$

To find the total potential at the origin, we integrate this expression over the entire hemispheric shell:

$$V = \int_0^{\pi/2} \int_0^{2\pi} \frac{k\rho_s a^2 \sin(\theta) d\theta d\phi}{a\sqrt{2(1-\cos(\theta))}}$$

Now, we can compute this double integral to find the potential $V$ at the origin.

Finished working Show work

The potential $V$ at the origin due to the charged hemispheric shell with uniform surface charge density $\rho_s$ is given by:

$$V = 2\sqrt{2}\pi a k \rho_s$$

Where:

- $k$ is Coulomb's constant, $k = \frac{1}{4\pi\epsilon_0}$
- $a$ is the radius of the hemispheric shell.

Substituting the value of $k$ in terms of $\epsilon_0$, the expression becomes:

$$V = \frac{2\sqrt{2}\pi a \rho_s}{4\pi\epsilon_0} = \frac{\sqrt{2}a\rho_s}{2\epsilon_0}$$

This is the potential at the origin due to the charged hemispheric shell.

Fig. 9: An instance where ChatGPT-4/CI seems to lack spatial awareness. The code is collapsed for brevity as it is not relevant to the intended subject of this figure. Here, on Problem 8 (calculating potential given a hemispheric shell of charge density), it did not realize that the distance from the center of a hemispherical shell to its surface is its radius. In this figure, the first response from ChatGPT-4/CI (zero-shot) is shown. However, ChatGPT-4/CI was able to catch its own mistake on being further asked to explain the “law of cosines” method used here, as shown in the next figure (Fig. 10)

ChatGPT-4/CI will sometimes catch its own mistakes:

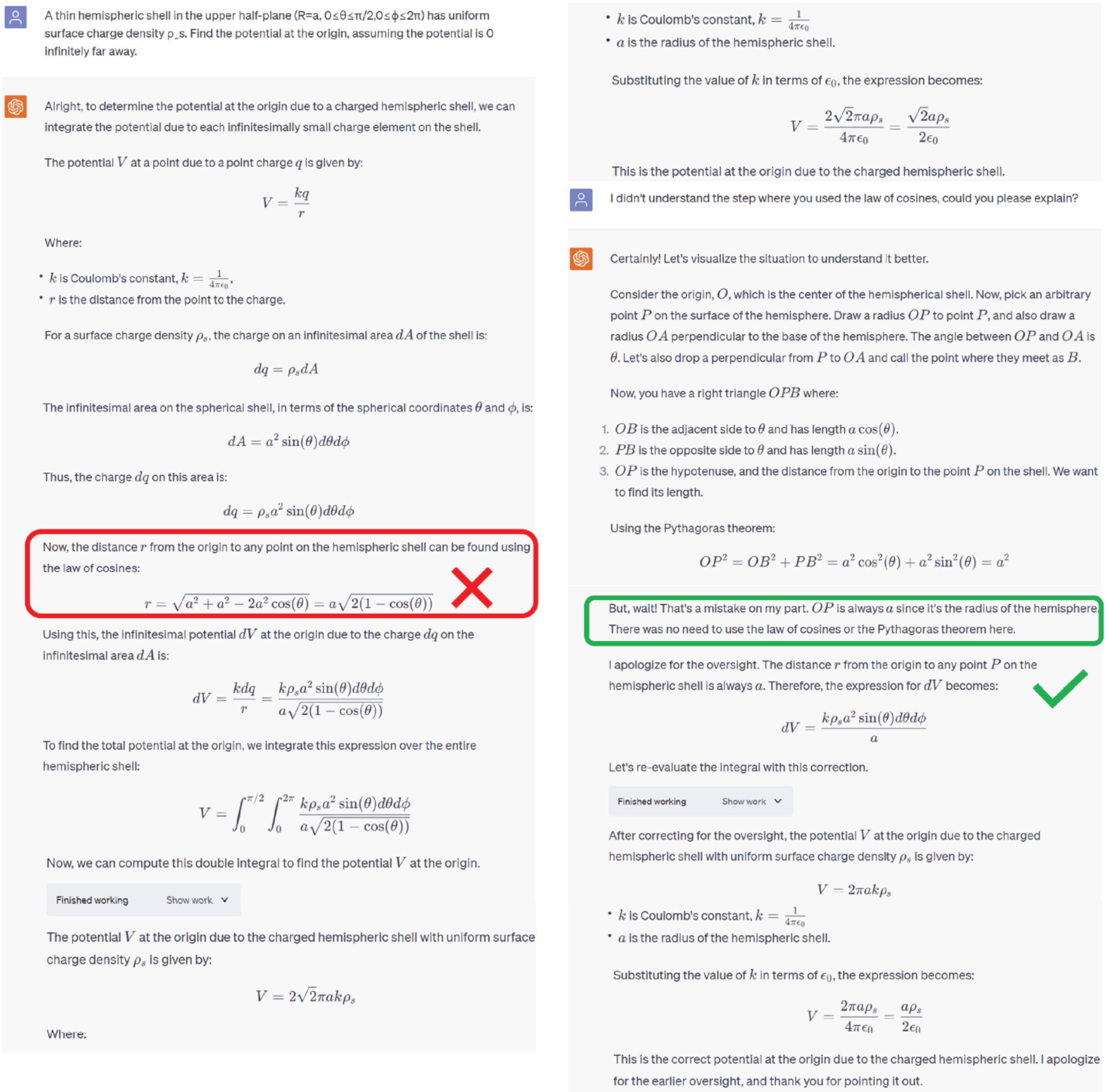

A thin hemispheric shell in the upper half-plane (R=a, 0≤θ≤π/2,0≤ϕ≤2π) has uniform surface charge density ρ_s. Find the potential at the origin, assuming the potential is 0 infinitely far away.

Alright, to determine the potential at the origin due to a charged hemispheric shell, we can integrate the potential due to each infinitesimally small charge element on the shell.

The potential $V$ at a point due to a point charge $q$ is given by:

$$V = \frac{kq}{r}$$

Where:

- $k$ is Coulomb's constant, $k = \frac{1}{4\pi\epsilon_0}$,
- $r$ is the distance from the point to the charge.

For a surface charge density $\rho_s$, the charge on an infinitesimal area $dA$ of the shell is:

$$dq = \rho_s dA$$

The infinitesimal area on the spherical shell, in terms of the spherical coordinates $\theta$ and $\phi$, is:

$$dA = a^2 \sin(\theta) d\theta d\phi$$

Thus, the charge $dq$ on this area is:

$$dq = \rho_s a^2 \sin(\theta) d\theta d\phi$$

Now, the distance $r$ from the origin to any point on the hemispheric shell can be found using the law of cosines:

$$r = \sqrt{a^2 + a^2 - 2a^2\cos(\theta)} = a\sqrt{2(1-\cos(\theta))}$$

Using this, the infinitesimal potential $dV$ at the origin due to the charge $dq$ on the infinitesimal area $dA$ is:

$$dV = \frac{kdq}{r} = \frac{k\rho_s a^2 \sin(\theta) d\theta d\phi}{a\sqrt{2(1-\cos(\theta))}}$$

To find the total potential at the origin, we integrate this expression over the entire hemispheric shell:

$$V = \int_0^{\pi/2} \int_0^{2\pi} \frac{k\rho_s a^2 \sin(\theta) d\theta d\phi}{a\sqrt{2(1-\cos(\theta))}}$$

Now, we can compute this double integral to find the potential $V$ at the origin.

Finished working Show work

The potential $V$ at the origin due to the charged hemispheric shell with uniform surface charge density $\rho_s$ is given by:

$$V = 2\sqrt{2}\pi a k \rho_s$$

Where:

- $k$ is Coulomb's constant, $k = \frac{1}{4\pi\epsilon_0}$
- $a$ is the radius of the hemispheric shell.

Substituting the value of $k$ in terms of $\epsilon_0$, the expression becomes:

$$V = \frac{2\sqrt{2}\pi a \rho_s}{4\pi\epsilon_0} = \frac{\sqrt{2} a \rho_s}{2\epsilon_0}$$

This is the potential at the origin due to the charged hemispheric shell.

I didn't understand the step where you used the law of cosines, could you please explain?

Certainly! Let's visualize the situation to understand it better.

Consider the origin, $O$, which is the center of the hemispherical shell. Now, pick an arbitrary point $P$ on the surface of the hemisphere. Draw a radius $OP$ to point $P$, and also draw a radius $OA$ perpendicular to the base of the hemisphere. The angle between $OP$ and $OA$ is $\theta$. Let's also drop a perpendicular from $P$ to $OA$ and call the point where they meet as $B$.

Now, you have a right triangle $OPB$ where:

1. $OB$ is the adjacent side to $\theta$ and has length $a\cos(\theta)$.
2. $PB$ is the opposite side to $\theta$ and has length $a\sin(\theta)$.
3. $OP$ is the hypotenuse, and the distance from the origin to the point $P$ on the shell. We want to find its length.

Using the Pythagoras theorem:

$$OP^2 = OB^2 + PB^2 = a^2\cos^2(\theta) + a^2\sin^2(\theta) = a^2$$

But, wait! That's a mistake on my part. $OP$ is always $a$ since it's the radius of the hemisphere. There was no need to use the law of cosines or the Pythagoras theorem here.

I apologize for the oversight. The distance $r$ from the origin to any point $P$ on the hemispheric shell is always $a$. Therefore, the expression for $dV$ becomes:

$$dV = \frac{k\rho_s a^2 \sin(\theta) d\theta d\phi}{a}$$

Let's re-evaluate the integral with this correction.

Finished working Show work

After correcting for the oversight, the potential $V$ at the origin due to the charged hemispheric shell with uniform surface charge density $\rho_s$ is given by:

$$V = 2\pi a k \rho_s$$

- $k$ is Coulomb's constant, $k = \frac{1}{4\pi\epsilon_0}$
- $a$ is the radius of the hemispheric shell.

Substituting the value of $k$ in terms of $\epsilon_0$, the expression becomes:

$$V = \frac{2\pi a \rho_s}{4\pi\epsilon_0} = \frac{a\rho_s}{2\epsilon_0}$$

This is the correct potential at the origin due to the charged hemispheric shell. I apologize for the earlier oversight, and thank you for pointing it out.

Fig. 10: First response and continuation of the chat shown in Fig. 9 (Problem 8) that shows how ChatGPT-4/CI can occasionally identify its own mistakes, though in our experiments this only occurred rarely, when ChatGPT-4/CI was asked to explain its choices. In this example, ChatGPT-4/CI realizes that the distance from the center of the hemisphere to its surface is the same as the radius $a$. In its first response, it used the law of cosines but did not evaluate that to find the distance to the surface of the hemisphere. The code is collapsed for brevity as it is not relevant to the intended subject of this figure.

Failure: ChatGPT, even ChatGPT-4/CI, will sometimes confidently defend wrong answers:

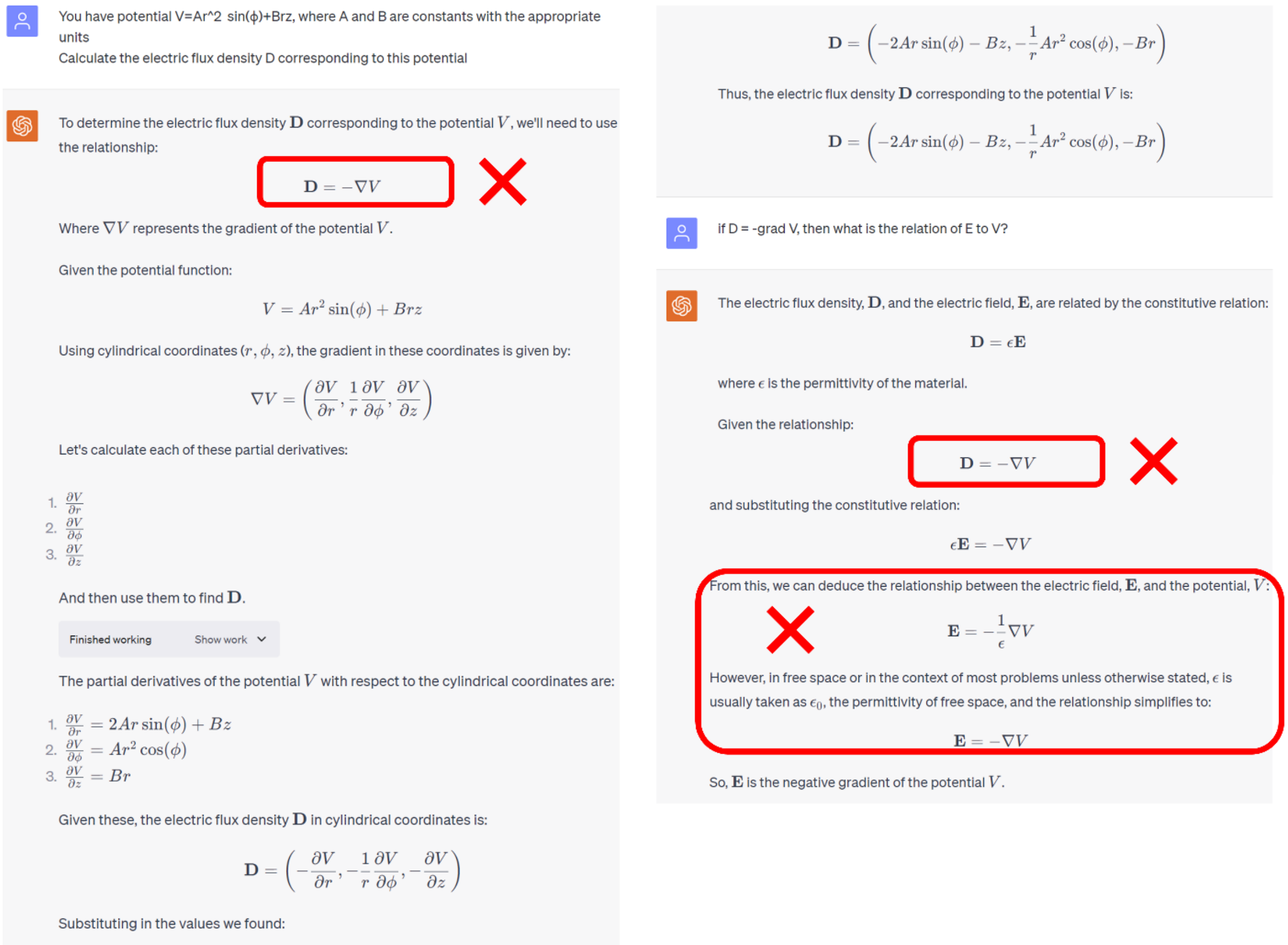

Fig. 11: An example where ChatGPT-4/CI confidently defends its wrong answer (it made a mistake in the relation between $D$ and $V$ in Problem 7, calculating the displacement from a potential in cylindrical coordinates). ChatGPT-3.5 is already prone to stating and sometimes defending wrong answers, but ChatGPT-4 or 4/CI was observed to double down on wrong answers (even though they were less common) with even more confidence.

ChatGPT-4/CI will sometimes make a mistake in stating the final answer even after working through it and plotting it correctly:

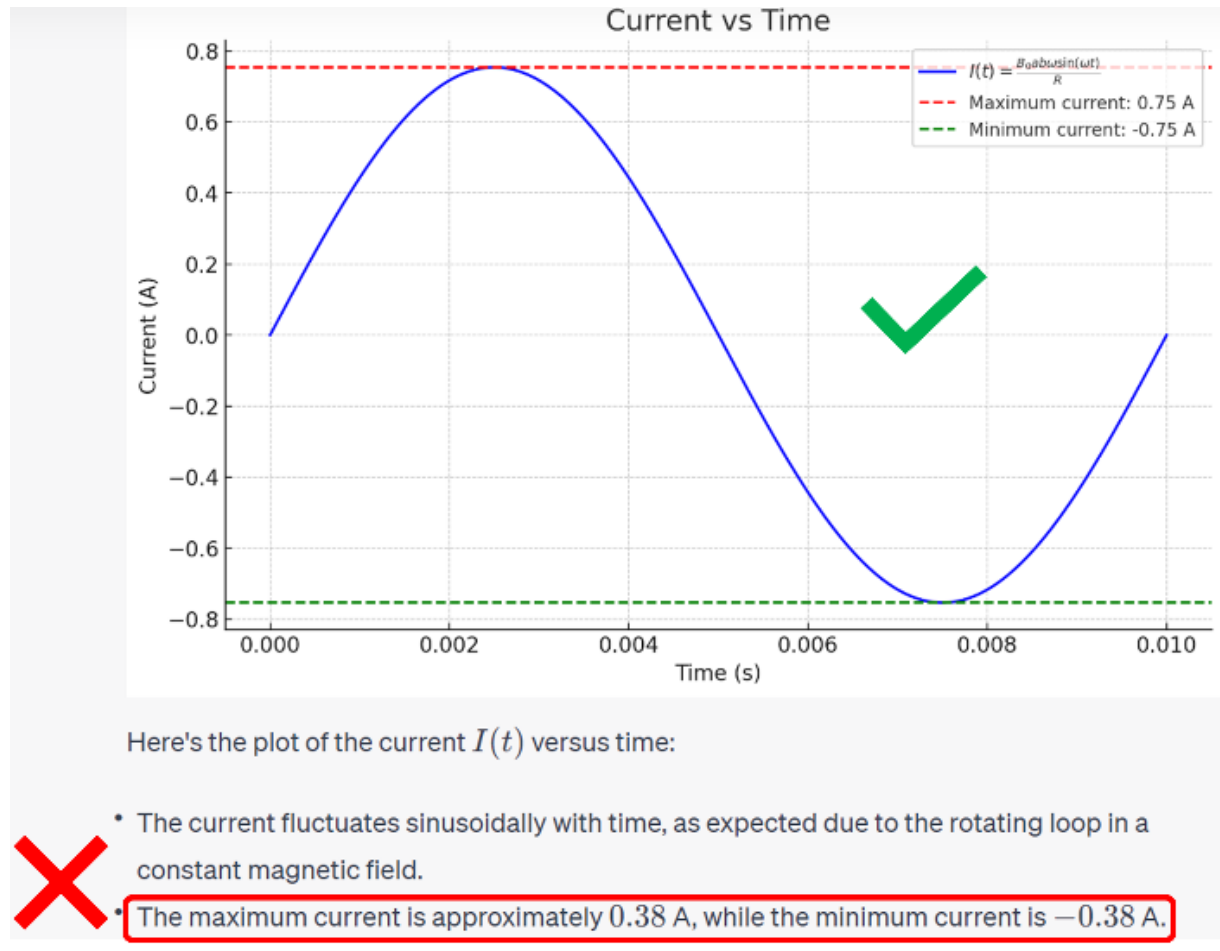

Fig. 12: An example of ChatGPT-4/CI working out Problem 13 correctly, (Only final plot is shown here; the response of ChatGPT-4/CI before this plot was the same as that shown in Fig. 1), but then misstating the values for maximum and minimum current (peak and trough values in the plot). We marked these cases as wrong in the evaluation scores.

### 5.2 Appendix: exact prompts

The following are the exact prompts given to ChatGPT for Problems 1-13:

1. Given matrices A and B, find (A-B)

   A=[−5 -5 −4;-1 -3 -4]

   B=[-9 -3 -10;-2 -7 -2]

2. Given A = [6.5 6.6 3.9 3.8; 0 6.0 7.9 3.1; 0 0 3.3 5.3; 0 0 0 3.0], find |A|

3. Calculate the total charge in a cube, 1 [meters] long each side located in the first octant (x≥0, y≥0, z≥0), with one corner at the origin. The volume charge density in the cube is rho_v = xy^2e^-3z mC/m^3 Start with writing down the charge density neatly first.

4. A line of electric charge with uniform density \rho = 3 \mu C/m exists along the z-axis between z=0 and z=9 [meters]. Find E in vector form at R(x, y, z) = (0, 22, 0) meters.

5. Calculate the divergence (strength of flux sources) of the vector field A = 2*sqrt(x^2 + y^2) x^ at the point (x,y,z)=(2,6,-7). Give three significant figures.

6. Calculate the divergence (strength of the flux sources) of the vector field A = rsin(2\phi) r^ + rcos(2\phi) \phi^ + (z^2) z^ at the point (r, \phi, z) = (19, 2pi/18, 1) Give three significant figures.

7. You have potential $V=Ar^2 \sin(\phi)+Brz$, where A and B are constants with the appropriate units Calculate the electric flux density D corresponding to this potential.

8. A thin hemispheric shell in the upper half-plane ($R=a$, $0\leq\theta\leq\pi/2, 0\leq\phi\leq 2\pi$) has uniform surface charge density $\rho_s$. Find the potential at the origin, assuming the potential is 0 infinitely far away.

9. Two small identical spheres have charges of +20 nC and -5 nC, respectively.
a. What is the force between them if they are separated by 10 cm?
b. The two spheres are brought into contact with each other (such that they equalize their charge due to conduction between the spheres) and then separated again by 10 cm. What is the force between them now?

10. You have a sphere of radius $a$, inside which the charge density can be written as $\rho_v = \rho_0/R^2$. Outside, the charge is zero. Find the electric field **E** outside the sphere.

11. The plane defined by $y=0$ separates two dielectrics. In $y<0$, we have medium 1, with permittivity $\epsilon_0$. In $y>0$, we have medium 2, with permittivity $\epsilon=\epsilon_r*\epsilon_0$. There is no charge on the interface. Just next to the interface, the field in medium 1 is $E_1=a\hat{x}+b\hat{y}$. Find $E_2$ in medium 2, just next to the interface.

12. A coaxial cable is made of two thin shells, each made of a perfect electric conductor. The inner shell radius is $r_1$, and the outer shell radius is $r_2$. In between the shells is air (you can assume it's free space). The length of the cable is $L\gg r_2$. a. (10 pts) Put charge +Q on the inner plate, and -Q on the outer plate. Calculate field everywhere in space (not just in between the plates!). b. (5 pts) Find the capacitance, C, of this configuration, based on the result in (a). Make sure to show your work step by step!

13. You have a rectangular conducting loop in three-dimensional space. One side of the loop is along the z axis with one end on the origin, with length b = 3 cm. Another side of the loop is in the x-y plane and has length a = 2 cm. The loop is a rectangle. This loop is rotating around the z axis at 6,000 RPM, and has internal resistance $R=0.5\ \Omega$. The orientation of the loop can be defined with an angle $\phi$ from the x axis. There is an external magnetic field which is constant, and pointing in the y direction everywhere, so $B=B_0\,\hat{y}$, where $B_0=1$ T.
a. What is the angular frequency, $\omega$, and what is the expression for $\phi(t)$ assuming that $\phi(t=0)=0$?
b. Find the expression for the magnetic flux, $\Phi(t)$, through the loop, in terms of $a,b,B_0,\omega$, and time t.
c. Find the expression for the current flowing through the loop. Plot the current flowing through the loop vs. time, I(t). On your plot, actually label the maximum and minimum current by plugging in all of the numbers.

### 5.3: Appendix: testing custom instructions

The following is a table on testing ChatGPT-4/CI with 3 custom instructions on 3 questions that it scored the lowest on without any custom instructions. The 3 custom-instruction statements are below the table.

| | Problem 4 | Problem 6 | Problem 7 |
|---|---|---|---|
| No custom prompt | 6/10 | 7/10 | 7/10 |
| Prompt 1 | 6/10 | 7/10 | 6/10 |
| Prompt 2 | 6/10* | 9/10 | 5/10 |
| Prompt 3 | 1/10[$] | 8/10 | 2/10[$] |

*ChatGPT-4/CI identifies its own mistake once in this question, and that solution was marked as correct. Prompt 2 required ChatGPT-4/CI to try the problem thrice, starting from scratch each time. Its solutions matched twice in this case and it reported that solution, which was correct. However, we only ever observed this once in our trials; in each of the other solutions, ChatGPT-4/CI's answers were the same in each of the three iterations (whether correct or wrong).

[$] In problems 4 and 7 with custom prompt 3, the text generation stopped before reaching the final answer in multiple instances, even if our instructions asked ChatGPT to work all the way through the problem In the table, the 10 trials for problems 4 and 7 each include such instances, wherein each time ChatGPT-4/CI stopped, we asked it to continue solving, which is arguably not a zero-shot run. However, even on discarding these instances that stopped and only counting truly zero-shot trials, ChatGPT-4/CI's performance on these problems with custom prompt 3 still looks poor.

Prompt 1:

> You will be solving problems in vector calculus and electrodynamics at an easy undergraduate level. For any problem below, we are using one of three coordinate systems:
>
> Cartesian with x, y, z
>
> Cylindrical with r, phi, and z
>
> Spherical with R, phi, and theta, where theta is the polar angle and phi is the same as the cylindrical phi.
>
> If something in the problem is ambiguous, note that.
>
> I recommend working out the problems analytically as much as possible, and then plugging in numbers at the end as needed.
>
> Solve the problem step by step. After each step, compare the expression with the expression in the previous step to make sure that you did not drop any terms or signs.
>
> ***PAY SPECIAL ATTENTION to all terms during differentiation, whether doing single variable differentiation or vector calculus***
>
> ***PAY SPECIAL ATTENTION while copying anything from your working to code***
>
> Acknowledge these custom instructions before attempting to solve the question.

Prompt 2: In addition to prompt 1 above, we add the following statements in the beginning of the chat (due to the word limit of the custom instruction field in the ChatGPT interface):

> You are a 3-stage iterative problem solver - for the given problem, solve it once, and use that answer as a hint for what the true, correct answer might be. Now using this answer as an input, solve the problem for the second time and find the answer. DO NOT assume that the method and formulas will be the same as before - start from scratch, rethinking the starting equation, deleting all code variables and not relying on previous code. Finally, using the answer from this second run as a more refined hint, solve the question a third time, using the given question and the latest solution as a hint. DO NOT assume that the method and formulas will be the same as before - start from scratch, rethinking the starting equation, deleting all code variables and not relying on previous code. Finally when all 3 iterations are done, compare your answers.

Prompt 3: In addition to prompt 1 above, we ask ChatGPT to imitate a student-teacher interaction at the beginning of the chat (again due to a word limit in the custom instruction field of the ChatGPT interface):

> Imitate a conversation between a student solving an electromagnetics question and a teacher checking the student's solution and making necessary corrections, if any.